\documentclass[letterpaper, 10 pt, conference]{ieeeconf}  

\IEEEoverridecommandlockouts                              

\usepackage{cite}
\usepackage{amsmath,amssymb,amsfonts}
\usepackage{algorithmic}
\usepackage{graphicx}
\usepackage[font=footnotesize]{subcaption}
\usepackage{textcomp}
\usepackage[table,xcdraw]{xcolor}
\usepackage{tabularx}
\usepackage{multirow}
\usepackage[hidelinks]{hyperref}
\usepackage{siunitx-v2}
\usepackage{setspace}
\usepackage[normalem]{ulem}
\usepackage{nicefrac}
\usepackage{pbalance}
\def\BibTeX{{\rm B\kern-.05em{\sc i\kern-.025em b}\kern-.08em
    T\kern-.1667em\lower.7ex\hbox{E}\kern-.125emX}}
\usepackage{aveasmargins}

\newcommand\subs[1]{_{\text{#1}}}

\renewcommand{\baselinestretch}{1}
\aveasSetMargins{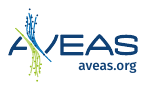}
\aveasSetIEEEFoot{2023}
\aveasSetIEEEHead{N. Neis and J. Beyerer, "A Two-Level Stochastic Model for the Lateral Movement of Vehicles Within Their Lane Under Homogeneous Traffic Conditions," 2023 IEEE 26th International Conference on Intelligent Transportation Systems (ITSC), Bilbao, Spain, 2023, pp. 1896-1903}{10.1109/ITSC57777.2023.10421975}

\title{\LARGE \bfseries {A Two-Level Stochastic Model for the Lateral Movement of Vehicles Within Their Lane Under Homogeneous Traffic Conditions}$^*$}

\author{Nicole Neis$^{1}$, 
Juergen Beyerer$^{2,3}$
\vspace{-9pt}
\thanks{*This publication was written in the context of the AVEAS research project (www.aveas.org), funded by the German Federal Ministry for Economic Affairs and Climate Action (BMWK) within the program ``New Vehicle and System Technologies''.}%
\thanks{
\raggedright
$^{1}$Porsche Engineering Group GmbH, 71287 Weissach, Germany,
{\tt\footnotesize nicole.neis@porsche-engineering.de}}%
\thanks{
\raggedright
$^{2}$Fraunhofer IOSB, 76131 Karlsruhe, Germany}%
\thanks{
\raggedright
$^{3}$Karlsruhe Institute of Technology (KIT), Vision and Fusion Laboratory (IES), 76131 Karlsruhe, Germany
}%
}

\begin{document}

\maketitle

\begin{abstract}
The lateral position of vehicles within their lane is a decisive factor for the range of vision of vehicle sensors. This, in turn, is crucial for a vehicle's ability to perceive its environment and gain a high situational awareness by processing the collected information. When aiming for increasing levels of vehicle autonomy, this situational awareness becomes more and more important. Thus, when validating an autonomous driving function the representativeness of the submicroscopic behavior such as the lateral offset has to be ensured. With simulations being an essential part of the validation of autonomous driving functions, models describing these phenomena are required. Possible applications are the enhancement of microscopic traffic simulations and the maneuver-based approach for scenario-based testing. This paper presents a two-level stochastic approach to model the lateral movement of vehicles within their lane during road-following maneuvers under homogeneous traffic conditions. A Markov model generates the coarse lateral offset profile. It is superposed with a noise model for the fine movements. Both models are set up using real-world data. The evaluation of the model shows promising qualitative and quantitative results, the potential for enhancements and extreme low computation times (\num{10000} times faster than real time).
\end{abstract}

\section{INTRODUCTION}\label{sec:Introduction}
\begin{figure*}
\centering
\begin{subfigure}{0.18\textwidth}
	\includegraphics[height=2.8cm]{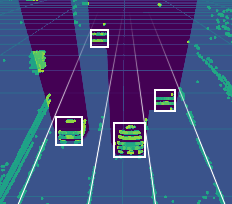}
	\caption{no offset}
	\label{fig:no_offset_lidar}
\end{subfigure}
\hfill
\begin{subfigure}{0.18\textwidth}
	\includegraphics[height=2.8cm]{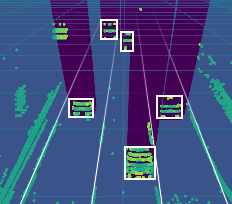}
	\caption{with offset }
	\label{fig:offset_lidar}
\end{subfigure}
\hfill
\begin{subfigure}{0.3\textwidth}
	\includegraphics[height=2.8cm]{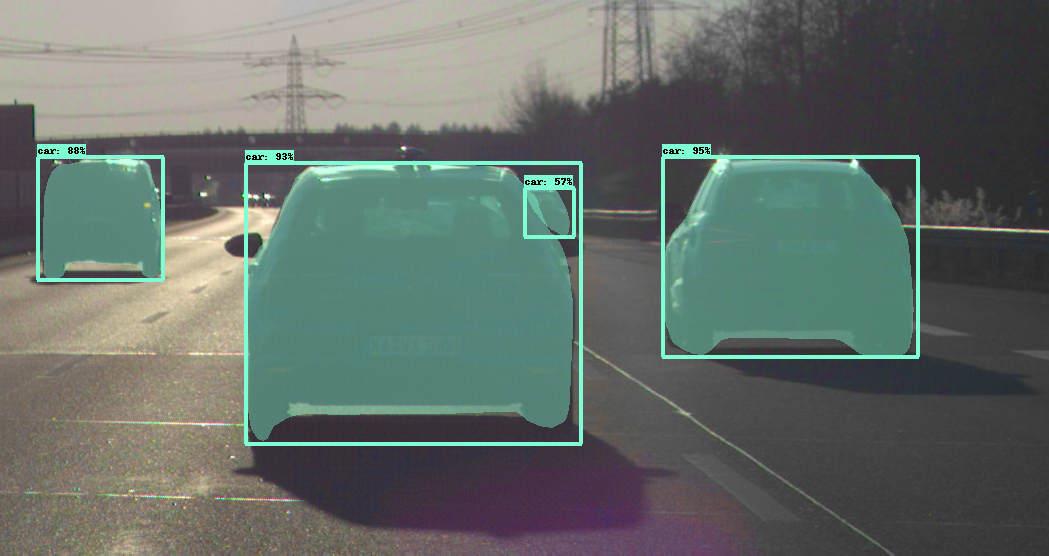}
	\caption{no offset }
	\label{fig:no_offset_camera}
\end{subfigure}
\hfill
\begin{subfigure}{0.3\textwidth}
	\includegraphics[height=2.8cm]{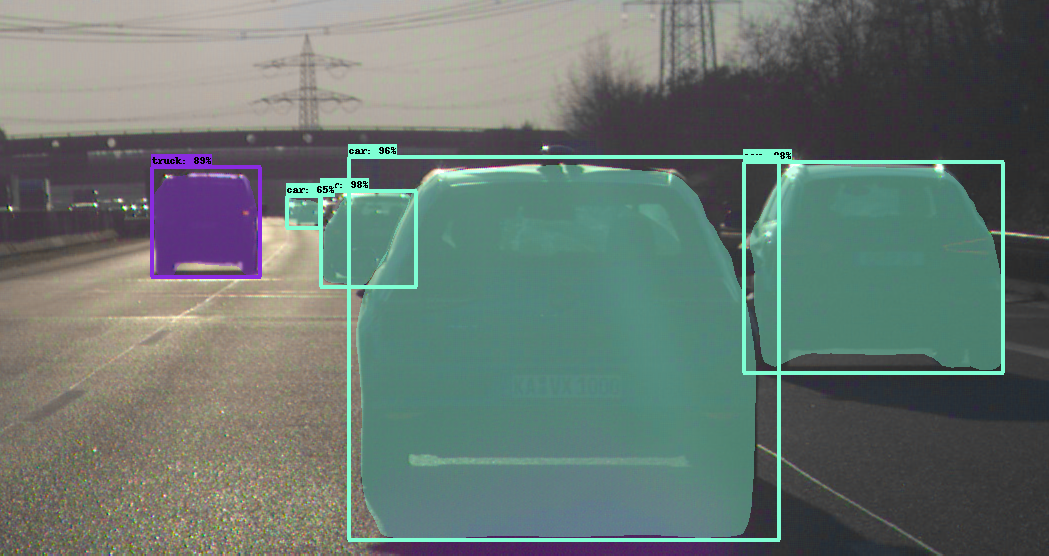}
	\caption{with offset }
	\label{fig:offset_camera}
\end{subfigure}
\caption{Effect of lateral offset between vehicles on same lane on the forward perception range for LiDAR (Valeo Scala Gen. 2, DBSCAN-based detection \cite{ester1996density}) and camera (Mask R-CNN / MS COCO \cite{he2017mask}). The ego lane perception range varies between \num{30}~\si{\metre} and \num{110}~\si{\metre} (LiDAR) and $\approx$ \num{200}~\si{\metre} (camera), motivating the goal to realistically represent this effect in virtual validation.}
\label{fig:range_of_vision_illustration}
\end{figure*}

When aiming for increasing a vehicle's level of autonomy, sensors such as cameras, LiDAR and radar take over the perception task. As in case of a human driver, their range of vision is highly relevant for the maximum degree of situational awareness they can achieve. 
As illustrated in Fig.~\ref{fig:range_of_vision_illustration}, the lateral offset between vehicles is decisive for the range of vision of vehicle sensors and the number of objects extracted out of the sensor data. Moreover, the lateral movement of vehicles within their lane is a challenging factor for an autonomous driving (AD) function such as a cut-in detection. For a reliable prediction of cut-ins, such a function needs to differentiate the characteristics of an upcoming cut-in from regular movement within a lane. Also methods aiming for a general anticipation of maneuvers based on real-world driving data collected from the vehicles interior and exterior to assist and warn the driver have been developed (see \cite{jainRecurrentNeuralNetworks2015}).  
For validating AD functions relying on the perception of the environment, the so called \emph{submicroscopic} behavior of the vehicles such as the lateral position within a lane has to be realistic with respect to real-world behavior. 
When testing under real-world conditions, this can be assumed to be fulfilled. However, as outlined in \cite{wachenfeldFreigabeAutonomenFahrens2015}, exclusively validating an AD function using classical methods such as field tests has become infeasible with increasing autonomy and growing Operational Design Domains. Thus, classical field tests are complemented by simulations. Options for this are microscopic traffic simulations \cite{krajzewiczTrafficSimulationSUMO2010, casasTrafficSimulationAimsun2010, sykesTrafficSimulationParamics2010, fellendorfMicroscopicTrafficFlow2010} and so called \emph{scenario}-based testing within simulations \cite{schuldtBeitragFuerMethodischen2016, elrofaiSCENARIOBASEDSAFETYVALIDATION2018, pfefferSzenariobasierteSimulationsgestuetzteFunktionale2020, putzSystemValidationHighly2017, riedmaierSurveyScenarioBasedSafety2020, sipplIdentificationRelevantTraffic2020}. For the latter, two scenario description approaches can be differentiated: the \emph{trajectory}-based and the \emph{maneuver}-based one \cite{neisLiteratureReviewOnManeuver2023}. When exactly replaying recorded trajectories, realistic submicroscopic behavior during scenario-based testing can be ensured. However, as explained in \cite{neisLiteratureReviewOnManeuver2023}, using the maneuver-based approach instead, brings the advantage of increasing the variety and number of scenarios by simple means and thus could be an enrichment for scenario-based testing. 
To ensure its validity, submicroscopic behavior models are needed. Other than approaches for trajectory prediction such as \cite{maResearchMultiobjectiveTrajectory2021}
aiming for an exact prediction of the upcoming short-term trajectory, submicroscopic behavior models have the goal to cover the space of all possible behaviors, even if they appear only rarely, for the full duration of a simulation. 
Also for microscopic traffic simulations further submicroscopic models are required to increase their realism especially for the lateral movement. The use case of a cut-in detection function demonstrates, how a lack of representativeness can yield simulation results not suitable for drawing reliable conclusions on an AD function's real-world behavior: if a simulation neglects the lateral position of a vehicle within its lane and simply locates it in the center, a function trained with this simulation would label every slight deviation from the center as an upcoming cut-in. While this would be correct in the simulation, the performance and safety of this function under real-world conditions are unclear.

After the introduction in the current section, an overview over related work is given in Sec.~\ref{sec:RelatedWork}. Section~\ref{sec:Dataset} describes the datasets used for calibrating the model that is introduced in Sec.~\ref{sec:Model}. The results obtained using this model are presented and discussed in Sec.~\ref{sec:Results}. Possible model enhancements are outlined in Sec.~\ref{sec:ModelEnhancement}. Section~\ref{sec:Conclusion} summarizes the key findings.

\section{RELATED WORK}\label{sec:RelatedWork}
As outlined in Sec.~\ref{sec:Introduction}, there is the need for submicroscopic behavior models describing the lateral movements of vehicles within their lane. 
Nevertheless, many microscopic traffic simulation tools such as SUMO \cite{krajzewiczTrafficSimulationSUMO2010}, Aimsun \cite{casasTrafficSimulationAimsun2010} and Paramics \cite{sykesTrafficSimulationParamics2010} concentrate on longitudinal movement and neglect lateral vehicle movement apart from lane changes. An exception is Vissim \cite{fellendorfMicroscopicTrafficFlow2010} where the so called \emph{strip-based} approach \cite{mathewStripbasedApproach2015} has been implemented. However, it is targeted at realistically modeling so called \emph{heterogeneous} traffic conditions characterized by poor lane discipline and a great variety of vehicles \cite{mallikarjunaHeterogeneousTrafficFlow2011} 
and only allows for discrete lateral movement. Thus, it is not suitable for the envisaged use case of homogeneous traffic conditions. 
A method suitable for homogeneous traffic conditions and allowing for continuous lateral movement is proposed in \cite{chakrobortyMicroscopicModelingDriver2004}: the lateral position of a vehicle is influenced by the superposition of potential fields emanated by scenery elements and traffic participants. A disadvantage of this model is that on an empty, uniform road, the vehicle will always drive in the center which is not representative for real-world traffic. Moreover, the method has not been calibrated with real-world driving data. A similar approach as in \cite{chakrobortyMicroscopicModelingDriver2004} is taken by Delpiano \cite{delpianoUnderstandingLateralDimension2021}. 
For calibrating his model, Delpiano uses the NGSIM dataset \cite{colyarUSHighway1012007, halkiasInterstate80Freeway2006}. However, he details problems introduced by this such as inaccuracies arising from the measurements and applied post-processing, affecting also the lateral position. 
Another limitation of data collected from static infrastructure sensors is that each detected vehicle is only observed for a short time. Thus, models on general vehicle behavior can be derived, however, 
it is not possible to capture the driving characteristics of individual vehicles and develop driver-specific behavior models. Though, this is of interest for simulations as the aim is to model not only average traffic conditions but to cover the full range, including also extreme cases such as a specific highly aggressive driver that might has been detected in reality. In \cite{qiStochasticLateralNoise2022}, lateral movement within a lane is included into microscopic behavior models using a stochastic differential equation. Its parameters are as well calibrated based on the NGSIM dataset. The model also enables continuous lateral movement and can reproduce real-world lateral offset distributions. However, these might be distorted for the same reasons as in \cite{delpianoUnderstandingLateralDimension2021} and show the same limitation of only being able to capture general but not individual vehicle behavior. The approach to avoid divergence by a mapping on the image space of the noise also suggests that, for longer simulation durations, the lateral behavior might degenerate towards extremal positions. Moreover, due to the complex approach of a stochastic differential equation, an extension of the model to consider the effects of static and dynamics elements of the environment (e.g. vehicles on a neighboring lane) on the lateral movement is expected to be difficult.  
Current maneuver-based approaches neglect \cite{heinzTrackScenariobasedTrajectory2017} or simplify the lateral offset behavior \cite{montanariManeuverbasedResimulationDriving2021}.

This paper presents a lightweight model for the lateral movement of a vehicle within its lane during road-following maneuvers under homogeneous traffic conditions. It allows for continuous lateral movement and is able to capture the characteristics of different tours used for calibrating the model. Every case, in which a vehicle is keeping its lane and driving with a longitudinal velocity of at least \num{40}~\si{\kilo\metre\per\hour} is considered a road-following maneuver, independent of other factors such as the acceleration, passing maneuvers of other vehicles, or a vehicle in front. The restriction of the longitudinal velocity is applied to exclude the effect of traffic jams from the model. More details on how these influence the lateral offset behavior of vehicles are given in Sec.~\ref{sec:Velocity}.

\section{DATASET}\label{sec:Dataset}
The calibration of the model presented in Sec.~\ref{sec:Model} is based on data collected during real-world test drives on German highways. The recording vehicle used for the two single-driver tours outlined below was a Porsche Cayenne equipped as described in \cite{haselbergerJUPITERROSBased2022}. The data used for the practical applications were obtained from the ego vehicle's bus signals with a frequency of about \num{25}~\si{\hertz}. The bus signals are processed data from the vehicle's control units. They include information on the ego vehicle's dynamics, indicator status, brake pressure, gear, etc. In particular, they provide information obtained from processing the series camera data. From this, object lists giving the longitudinal and lateral distances, relative velocities and class of surrounding traffic participants as well as the distance of the ego vehicle to the left and right lane marking are obtained.
To filter out erroneous measurements, down-sampling to a frequency of \num{5}~\si{\hertz} is applied. It is assumed that below this frequency threshold significant behavioral features can likely be determined and thus there is no loss of human driving behavior caused by this down-sampling. 
At the same time, it implies that the scope of the introduced model is the description of human driving behavior and not the driving physics.  

In Sec.~\ref{sec:Model} and \ref{sec:Results}, the model calibration and results are exemplified on the basis of a recording drive of a single driver. It had a duration of about \num{1.5}~\si{\hour} and a length of approximately \num{170}~\si{\kilo\metre} (\emph{first tour dataset}). After removing lane changes and traffic jam situations, about \num{50}~\si{\min} of driving data distributed over the full \num{1.5}~\si{\hour} remained that are used for setting up the model.
As reference, data obtained from another tour by a different driver are used at a later stage. This second drive had a duration of about \num{1}~\si{\hour} and covered approximately \num{106}~\si{\kilo\metre} (\emph{second tour dataset}). Also in this case, about \num{50}~\si{\min} remained for usage. In both cases, the two signals giving the distance to the left and the right lane marking were used to derive the lateral position of the ego vehicle within its lane. This is the only input needed for calibrating the presented model. 
To be able to compare the lateral offset behavior even if the lanes have different widths, the relative lateral position within the lane is used in the following. A relative lateral position of \num{0} indicates that the center of the vehicle is exactly in the middle of the lane, \num{-0.5} indicates that the center of the vehicle is exactly over the left lane marking, if the relative lateral position is equal to \num{0.5} the center of the vehicle is over the right lane marking. 

To complement the two single-driver datasets, the so called \emph{extended dataset} is used in addition. It consists of several recording drives of different drivers on German highways with a total duration of \num{15}~\si{\hour} and a length of about \num{1600}~\si{\kilo \metre}. Besides the Porsche Cayenne, a Porsche Taycan delivering the same bus signals was used for recording. The recording frequencies lie between \num{18} and \num{25}~\si{\hertz}. The extended dataset is used to derive possible advancements of the model (see Sec.~\ref{sec:ModelEnhancement}) and to make a reasonable selection of the data used for setting up the two-level stochastic model (see Sec.~\ref{sec:Velocity}).

\subsection{Velocity dependence of lateral movement}\label{sec:Velocity}
\begin{figure*}
\centering
\begin{subfigure}{0.3\textwidth}
	\includegraphics[width=\textwidth]{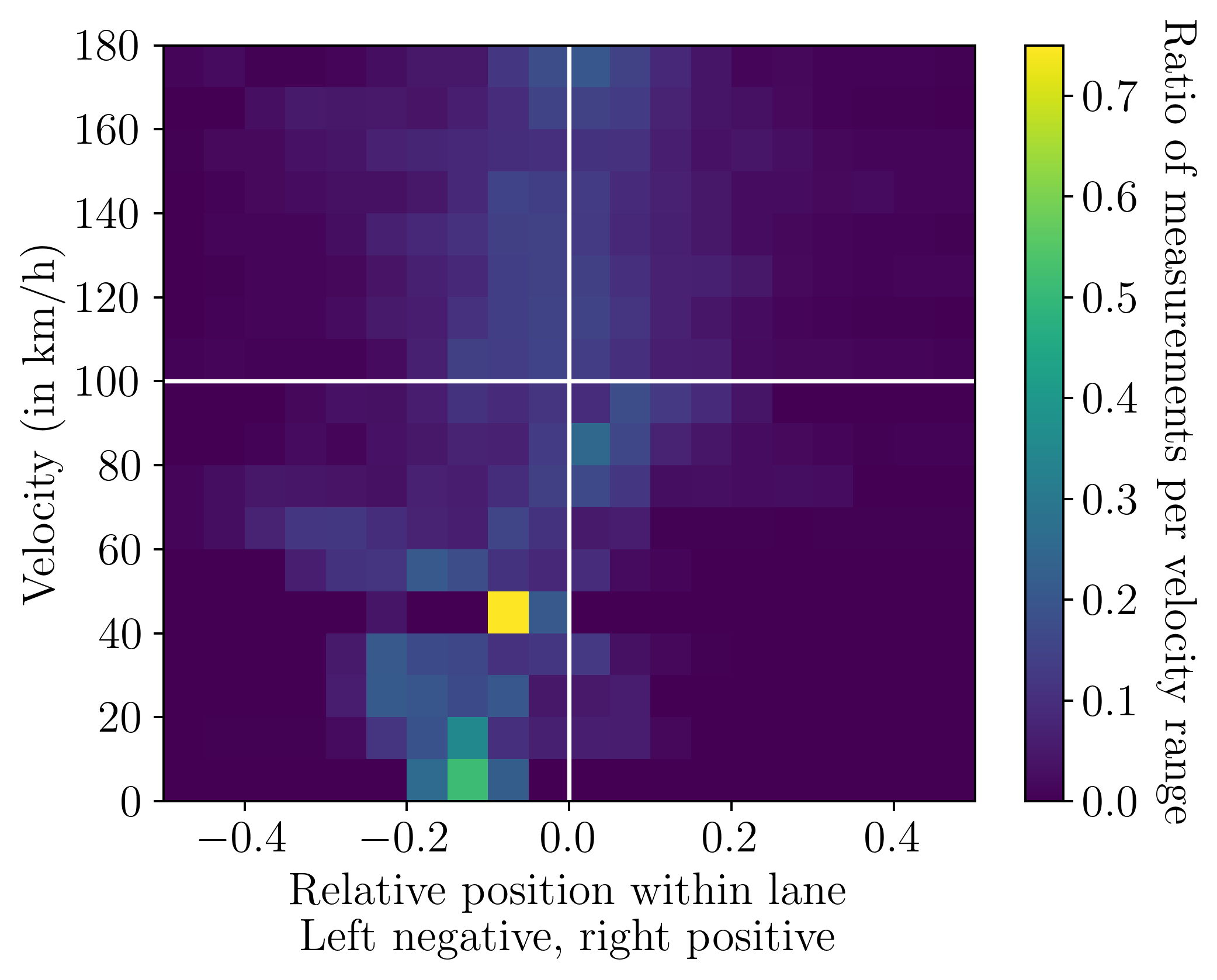}
	\caption{\centering left lane}
	\label{fig:lane_comparison_vehicles_left_velo}
\end{subfigure}
\begin{subfigure}{0.3\textwidth}
	\includegraphics[width=\textwidth]{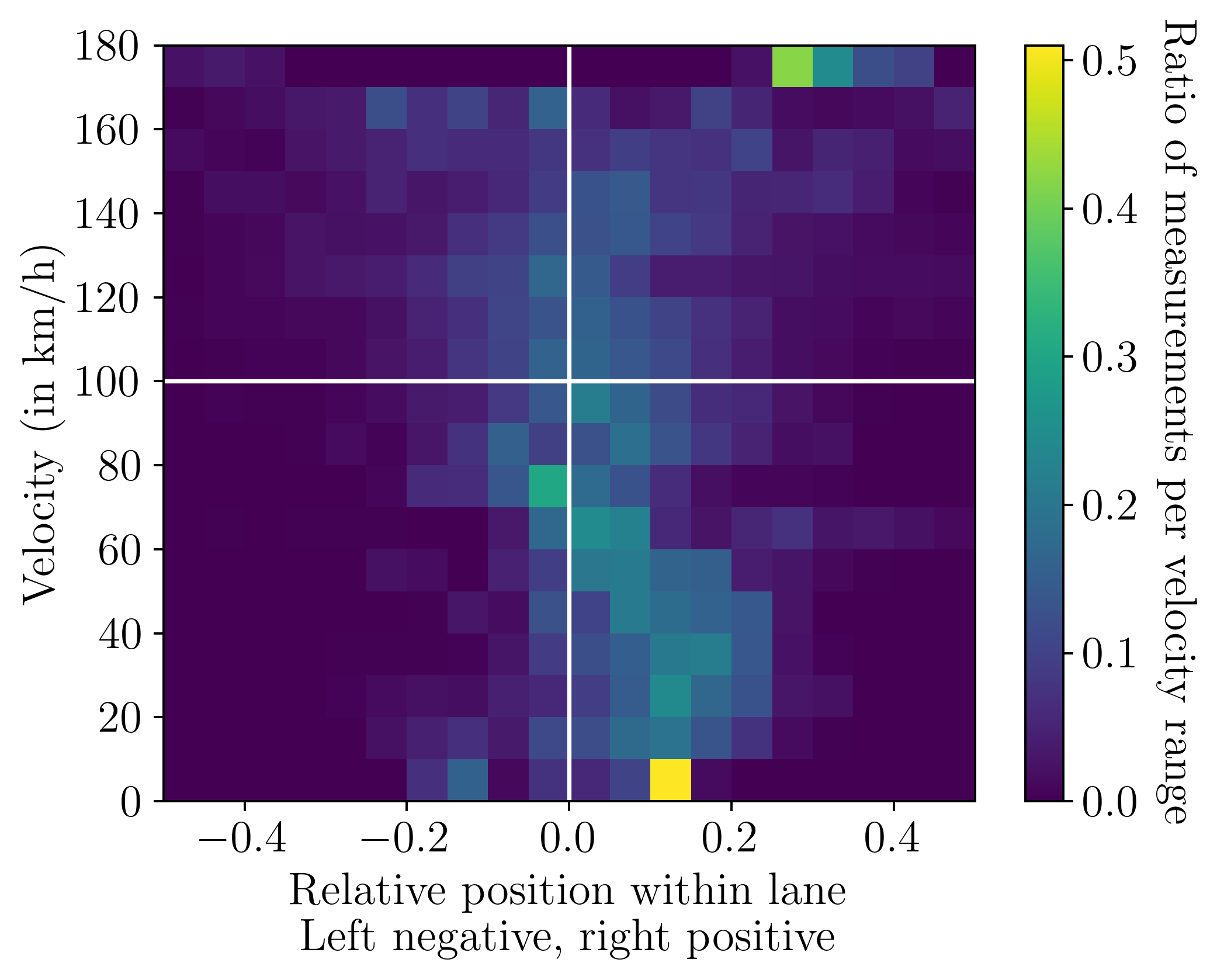}
	\caption{\centering center lane}
	\label{fig:lane_comparison_vehicles_center_velo}
\end{subfigure}
\begin{subfigure}{0.3\textwidth}
	\includegraphics[width=\textwidth]{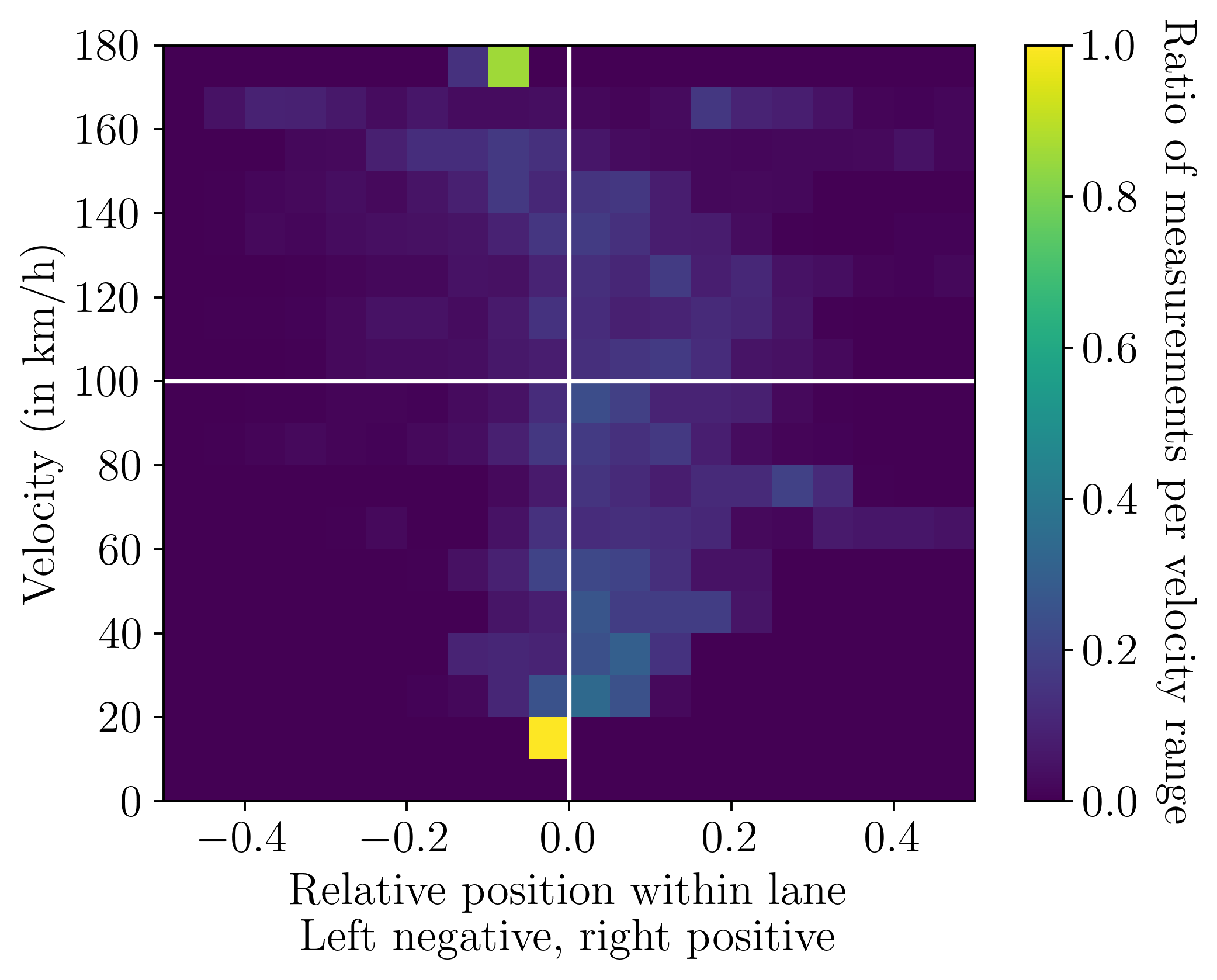}
	\caption{\centering right lane}
	\label{fig:lane_comparison_vehicle_right_velo}
\end{subfigure}
\caption{Lateral offset for different velocity ranges and lanes.}
\label{fig:velocity}
\end{figure*}

Within this section, the lateral offset behavior of vehicles in dependence of their velocity is put into focus. Fig.~\ref{fig:velocity} shows the relative lateral position of passenger cars on a left, right or center lane for different longitudinal velocities. Besides the bus signals giving the distance to the left and right lane marking, the object lists describing the surrounding traffic were used. Depicted are the measurements taken from all \emph{lead vehicles} (the vehicle in front of the ego vehicle) with a distance of not more than \num{100}~\si{\metre} from the ego vehicle contained within the extended dataset. The lead vehicles are used instead of the ego vehicles to avoid over-representation of certain ego vehicle drivers. Data analysis revealed that for higher distances between the ego vehicle and the lead vehicle inaccuracies occur, e.g. due to road curvature, therefore the distance restriction of \num{100}~\si{\metre} is introduced. 
Particularly interesting in Fig.~\ref{fig:velocity} are the low velocities: vehicles tend to be located further left when driving on the left lane and further right when driving on the center lane. This positioning within the lane reflects the creation of an emergency lane in case of a traffic jam. It motivates the restriction of the data used for setting up the two-level stochastic model to those for which the ego vehicle's longitudinal velocity was at least \num{40}~\si{\kilo\metre\per\hour}. For higher velocities, the size of the velocity- and lane-dependent effects seem to be smaller. Therefore, for the initial model presented in this paper, these effects are neglected. However, as interrelations between the longitudinal velocity and the lane with the lateral positioning can still be noted, these are to be considered in future models. More details on this are given in Sec.~\ref{sec:ModelEnhancement}.

\section{MODEL}\label{sec:Model}
\begin{figure*}
\includegraphics[width=\textwidth]{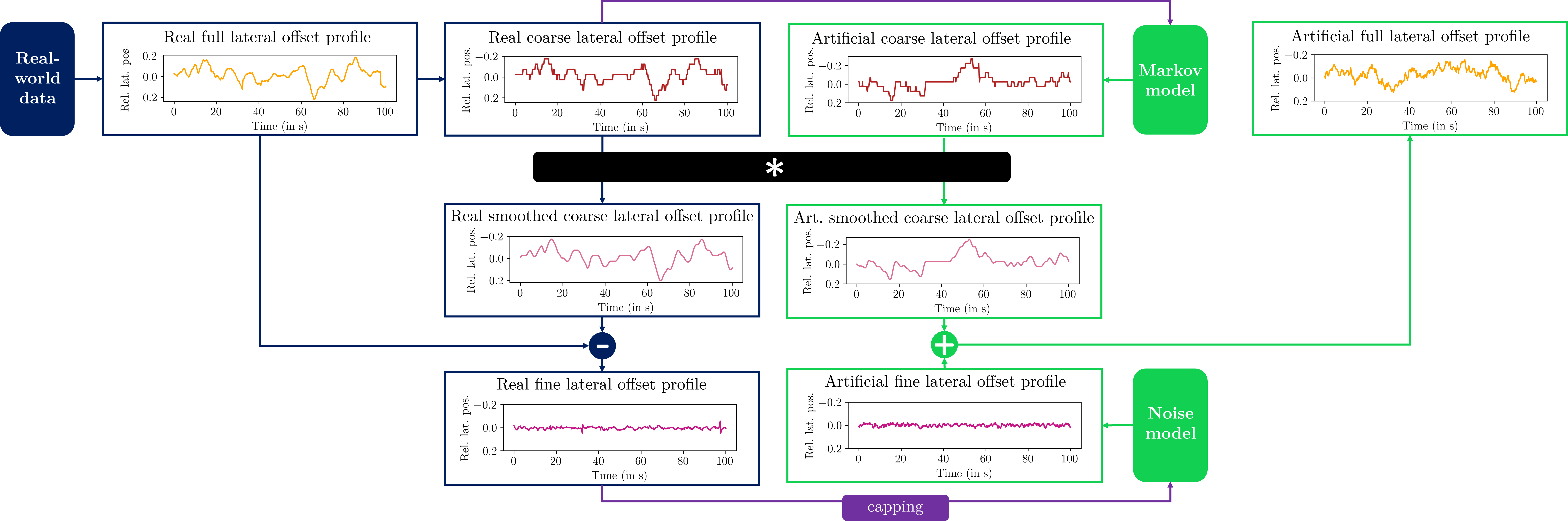}   
\caption{Illustration of the model setup (in blue, left side) and its inversion, the model usage (in green, right side). The $*$ operator stands for a convolution of the lateral offset profile with a Gaussian kernel. Elements in purple illustrate how the real data are used for calibrating the two stochastic models.}
\label{fig:modelWorkflow}
\end{figure*}

The core idea of the introduced model is the separation of the lateral movement of a vehicle into two independent parts: the \emph{coarse} movement $\kappa(i)$ and the \emph{fine} movement $\phi(i)$ over time steps $i$. If superimposed, they result in the observed lateral driving behavior of a vehicle within its lane
\begin{equation}
    x(i) = \kappa(i) + \phi(i)\;, \label{eq:decomposition}
\end{equation}
where the \emph{coarse} model $\kappa$ provides position-dependent behavior based on discrete states, to represent systematic behavioral properties in the lateral position, while the \emph{fine} model $\phi$ contributes residual, continuous behavioral properties, defined as being stochastically independent from the coarse behavior.
Both models are described in detail in the following sections. Fig.~\ref{fig:modelWorkflow} illustrates how the real offset profile is processed to set up the two-level stochastic model and how an inversion of this process -- the usage of the model -- generates an artificial lateral offset profile.

\subsection{Coarse movement: Markov model}
\begin{figure}
\includegraphics[width=0.5\textwidth]{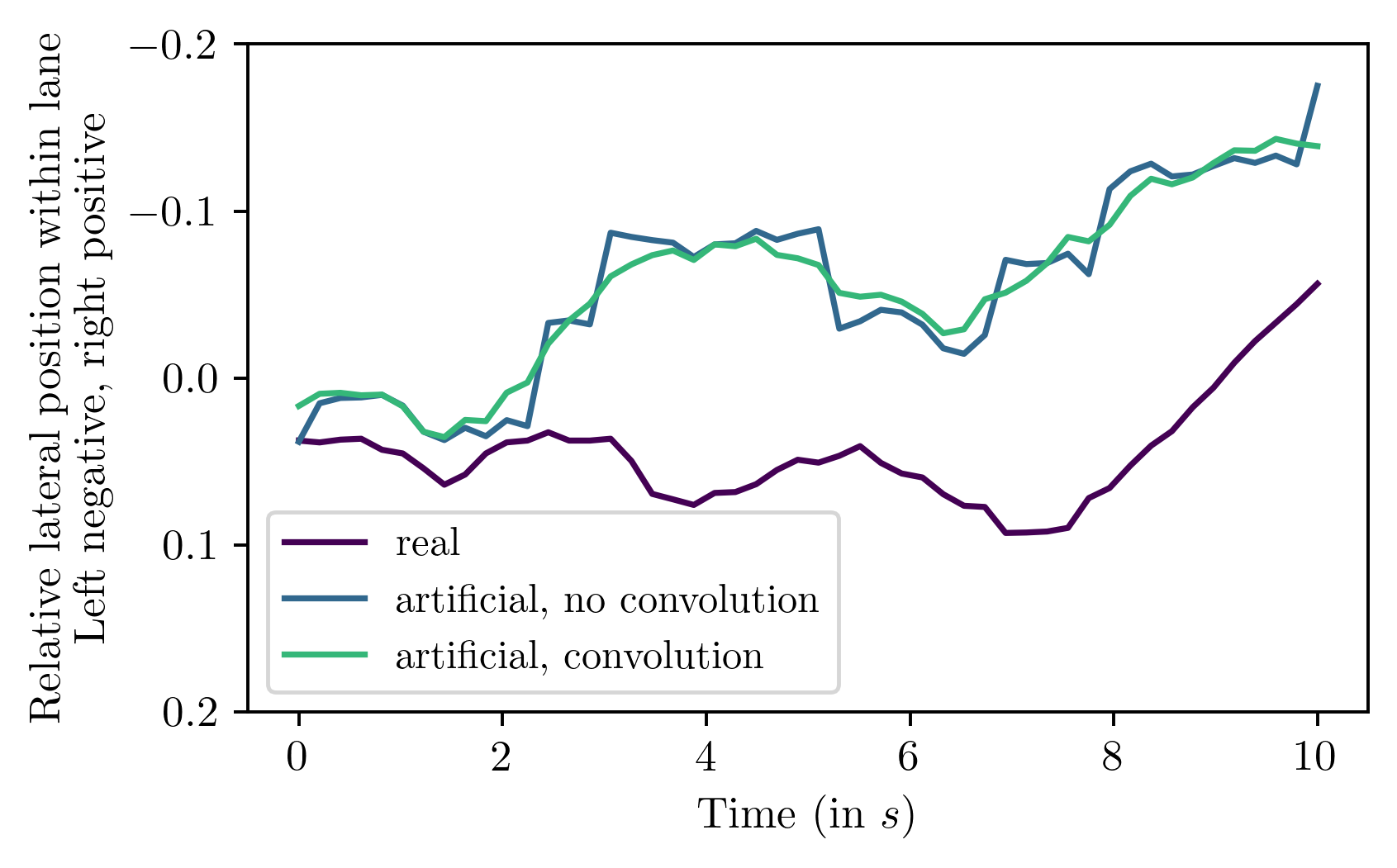} 
\caption{Illustration of the effect of convolving the coarse lateral offset profile generated by the Markov model with a kernel $g_s$, \mbox{$s = \num{0.6}~\si{\second}$} on the final artificial trajectories: clearly visible steps that are not observable in the real lateral offset profiles vanish.}
\label{fig:conv_illustration}
\end{figure}

To model the state-dependent, state-discrete dynamic behavior of $\kappa(i)$ stochastically, we use a Markov chain. This decision is motivated by the temporal continuity of a vehicle's lateral movement, making a position close to the current one more likely for the next time step than those further away.
We define the Markov chain over the space of lateral lane coordinates between \num{-0.5} and \num{0.5} by dividing it into $n_c$ segments of width $\nicefrac{1}{n_c}$, where we use a value of $n_c = 20$ in all practical applications. State transitions occur over time indices $i$. For the practical applications, we choose a time step of \num{0.2}~\si{\second}. The resulting transition matrix $T$ is of dimension $n_c \times n_c$. We parameterize it through the relative lateral position information from the first tour dataset. Thus, from the Markov model, we receive discrete lateral positions ranging from \mbox{$\num{-0.5}+\nicefrac{1}{2n_c}$} to \mbox{$\num{0.5}~-~\nicefrac{1}{2n_c}$}, resulting in a step function as lateral offset profile. We note that the direct outputs of the Markov chain, which we will denote as $\hat\kappa$, are still unsuitable for the model. By our model definition according to \eqref{eq:decomposition}, the remainder to achieve the complete lateral motion, the \emph{fine} movement $\phi$, should be stochastically independent from the \emph{coarse} movement. Hence, an independent $\phi$ cannot compensate sharp state transitions in $\kappa$ \emph{when they occur}. Instead, we must assure that $x(i)-\kappa(i)$ be, in good approximation, homogeneous over $i$. Therefore we let
\begin{equation}
    \kappa = \hat\kappa * g_s
\end{equation}
where $g$ is a Gaussian kernel with mean zero and standard deviation $s$ \mbox{($s = \num{0.6}~\si{\second}$} and support $\pm \num{1} \si{\second}$ in all practical applications). The choice of $s$ directly interacts with the obtained results for $\phi$ according to Sec.~\ref{sec:NoiseModel}: when chosen too sharp, discretization errors will dominate in $\phi$; when chosen too wide, the state-dependent results of $\hat\kappa$ will be erased, leaving the weaker model $\phi$ to primarily determine the lateral behavior. Fig.~\ref{fig:conv_illustration} illustrates the results obtained when choosing \mbox{$s = \num{0.6}~\si{\second}$} versus not applying any convolution with a kernel~$g$.

\subsection{Fine movement: noise model}\label{sec:NoiseModel}
\begin{figure}
\includegraphics[width=0.5\textwidth]{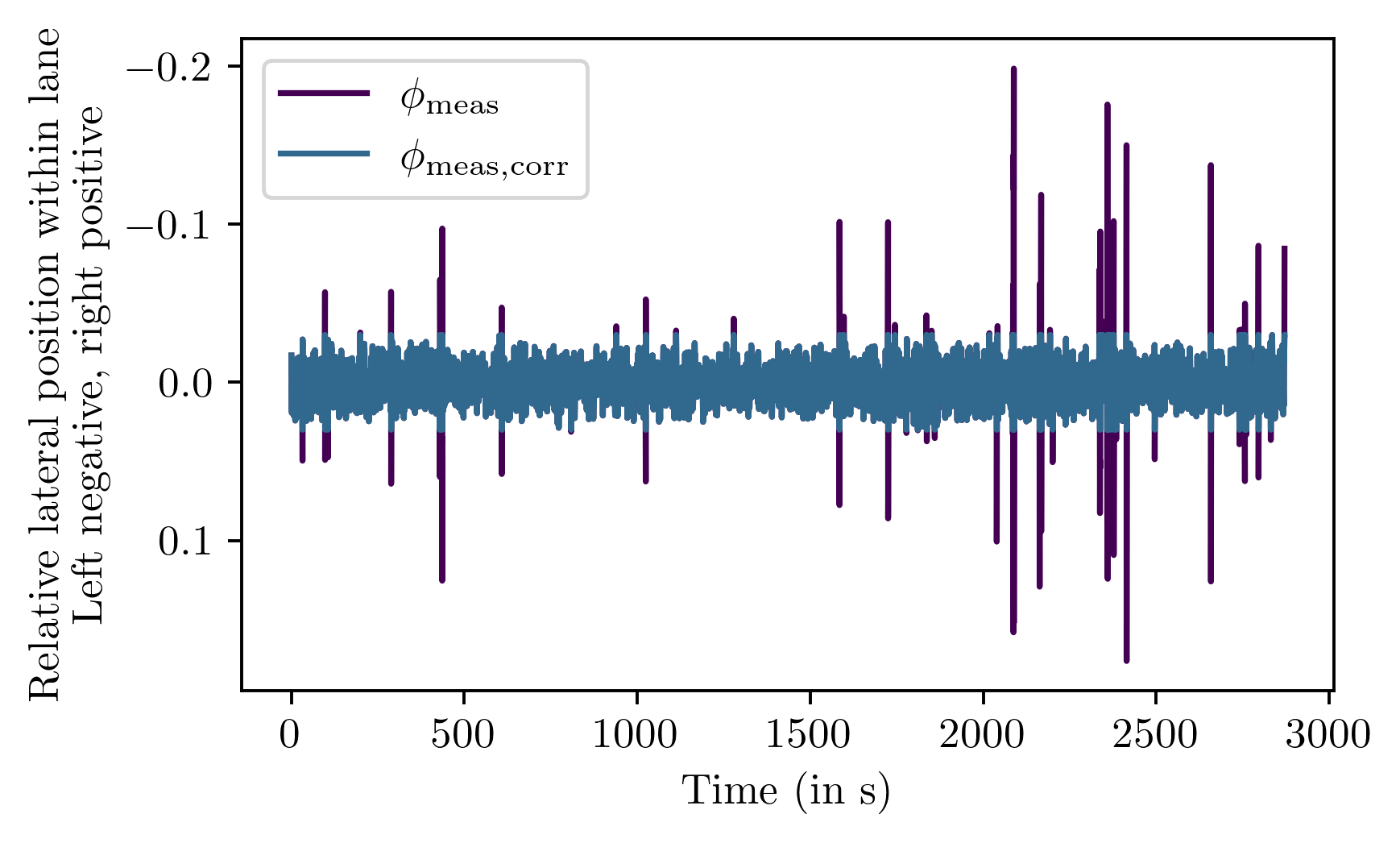} 
\caption{Capping of fine movement signal to avoid deviations due to peaks.}
\label{fig:corr_illustration}
\end{figure}

As initially stated, we assume the remaining fine movement $\phi$ to be independent of $\kappa$. The validity of this assumption is discussed in Sec.~\ref{sec:Results}. The fine movement of the driver is extracted out of measured data $x\subs{meas}$ in the following way (see Fig.~\ref{fig:modelWorkflow} for an illustration): 
\begin{equation}
    \phi\subs{meas} = x\subs{meas} - \hat\kappa\subs{meas} * g_s
\end{equation}
where $\hat\kappa\subs{meas}$ is given by $x\subs{meas}$ rounded to steps of $\nicefrac{2}{n_c}$.
The resulting $\phi_\mathrm{meas}$ is illustrated in Fig.~\ref{fig:corr_illustration}. As it can be seen, a few peaks occur in this signal. These peaks only make sense in combination with the coarse lateral position at the point of time they occur in the real driving data. Thus, they introduce a dependency between the fine and the coarse lateral movement. Consequently, if using $\phi_\mathrm{meas}$ for parametrizing $\phi$, stochastic independence between the coarse and the fine movement is not given, causing deviations. For that reason, $\phi_\mathrm{meas,corr}$ is used instead for parametrizing $\phi$. It is derived from $\phi_\mathrm{meas}$ by capping $\phi_\mathrm{meas}$ above and below a certain threshold $t_\phi^{\pm}$ (we choose $t^{\pm}_\phi = \pm 0.03$ in all practical applications).

\begin{figure}
\includegraphics[width=0.5\textwidth]{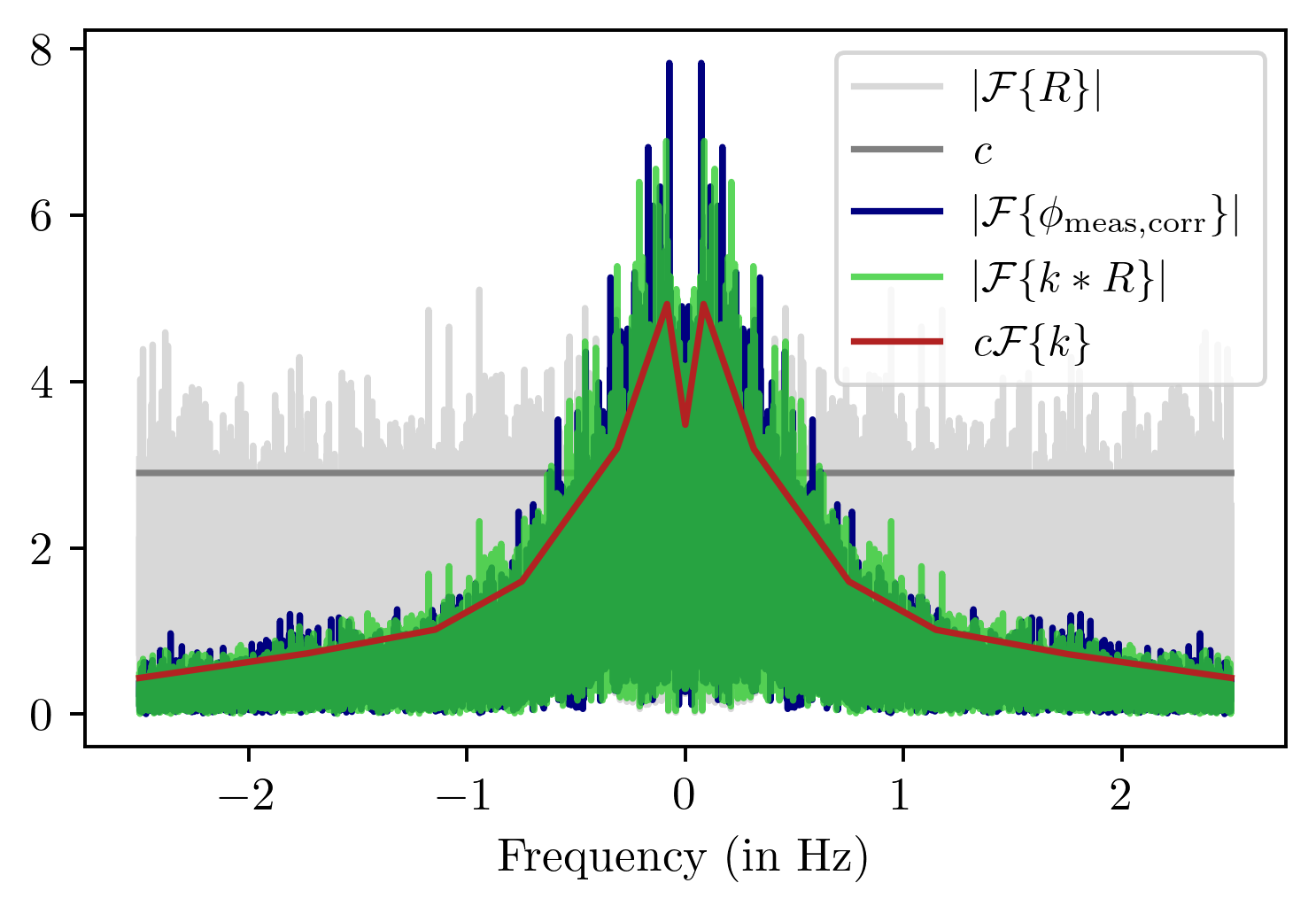}   
\caption{Illustration of kernel selection in frequency space. The Fourier transform of white noise is approximated by the constant value $c$.}
\label{fig:fft_fitting}
\end{figure}

As according to our model, $\phi$ is stochastic, independent of $\kappa$ and homogeneous across its value range, we can specify it through a noise model of unknown frequency composition w.r.t. constrained, uniform white noise $R(i)$. This allows us to compute
\begin{equation}
    \phi = R * k\label{eq:fine-movement}
\end{equation}
where $k$ is an unknown convolution kernel. In frequency space, $\mathcal{F}\{k\}$ acts as a real-valued damping function. Thus, we consider the Fourier-transform $\mathcal{F}\{\phi\subs{meas,corr}\}$ of the capped measured fine movement with respect to the Fourier-transform of the constrained uniform white noise $\mathcal{F}\{R\}$, and approximate $\mathcal{F}\{k\}$ as a piecewise linear function, as shown in Fig.~\ref{fig:fft_fitting}. $\mathcal{F}\{k\}$ is varied until $\mathcal{F}\{\phi\subs{meas,corr}\}$ and $\mathcal{F}\{k * R\}$ are in good accordance. Its inverse Fourier transform then provides $k$ for applying \eqref{eq:fine-movement}. The model for the fine movement can also be interpreted as sampling from a Kalman filter with a transition matrix reproducing a convolution with the kernel $k$ and the remaining terms are set to zero.

\section{RESULTS AND DISCUSSION}\label{sec:Results}
The developed model is evaluated using ten metrics applied to lateral offset snippets of \num{10}~\si{\second} duration. Let $X = [x_0, ..., x_m] \in \mathbb{R}^{m+1}$ with $x_i$ the relative lateral position at $i$ time steps after the initial time $t_0$ be such a lateral offset snippet. Then the used metrics are the following: $x_\mathrm{max}$ (maximum of $X$), $x_\mathrm{min}$ (minimum of $X$), $\bar{x}$ (mean of $X$), $\sigma$ (standard deviation of $X$), $x_\mathrm{0.5}$ (median of $X$), $x_\mathrm{0.25}$ (\num{25}~\% percentile of $X$), 
$x_\mathrm{0.75}$ (\num{75}~\% percentile of $X$), $r$ defined as $r := \vert x_\mathrm{max} - x_\mathrm{min}\vert$ (range of $X$), $\bar{x}_\mathrm{diff,10}$ (mean difference between two consecutive values in $X$ times~\num{10}), 
$\sigma_\mathrm{diff,10}$ (standard deviation of the difference between two consecutive values in $X$ times~\num{10}).
The metrics applied on $X$ directly indicate whether the same (extremal/mean) lateral positions are reached equally often in the real and artificial lateral offset snippets and whether these snippets have equal lateral ranges. The metrics applied on the difference of two consecutive values in $x$ allow to evaluate the temporal evolution of the artificial offset snippets and their similarity to the real ones.

To assess not only the full two-level stochastic model but also its components separately, the evaluation is split into four steps. For every evaluation, the first tour dataset is divided into snippets of \num{10}~\si{\second} duration. To compare the metrics for the real and the artificial lateral offset profiles, for each real lateral offset snippet, an artificial one having the same duration and initial lateral offset is generated.

\subsection{General model validity}
\begin{figure}
\includegraphics[width=0.5\textwidth]{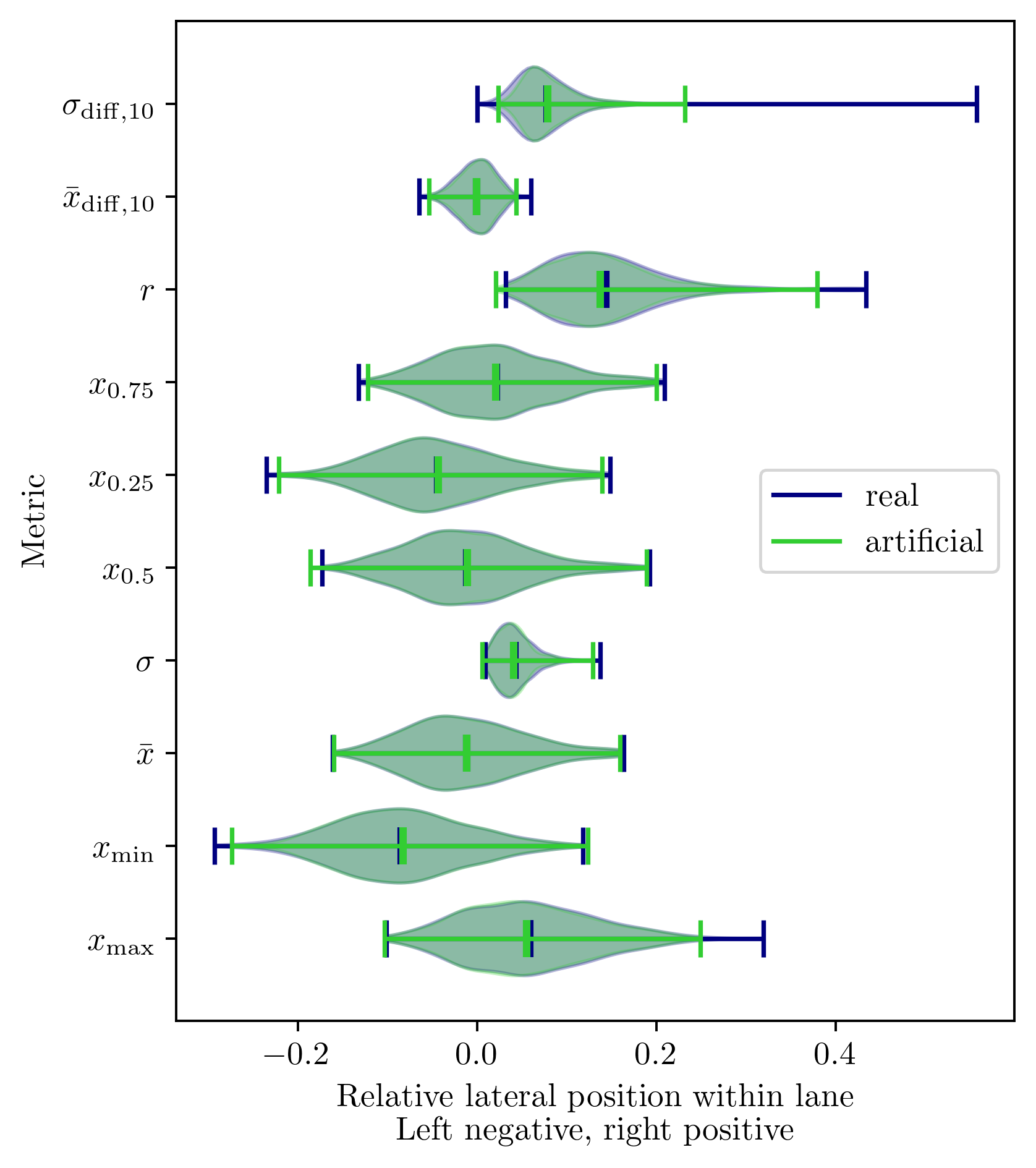} 
\caption{Comparison of metrics for real lateral offset profiles and lateral offset profiles generated by keeping the real coarse movement and adding the time-shifted capped real fine movement.}
\label{fig:shift}
\end{figure}

First, the general validity of separating the lateral movement of a vehicle into a coarse and a fine movement is discussed. For this, artificial lateral offset profiles are generated in the following way: the real coarse lateral offset profile is kept and superposed with the capped real fine movement which is circularly shifted by \num{5}~\si{\second}. Fig.~\ref{fig:shift} shows the results obtained when evaluating the defined metrics. In the here depicted as well as all following violin plots, the vertical lines refer to the minimum, mean and maximum value. In Fig.~\ref{fig:shift} it can be seen that the metrics differ only slightly from the ones of the real lateral offset snippets. Thus, shifting the fine part does not substantially change the characteristics of a driver's lateral movement for the chosen model set up. This confirms the assumption made in Sec.~\ref{sec:NoiseModel} and shows that indeed in our practical application the fine and the coarse lateral movement can be considered independent and the separation can be rated permissible. At the same time, Fig.~\ref{fig:shift} also indicates the maximum agreement with the developed model one can achieve when generating an artificial lateral offset profile.  

\subsection{Evaluation of Markov model}
\begin{figure}
\includegraphics[width=0.5\textwidth]{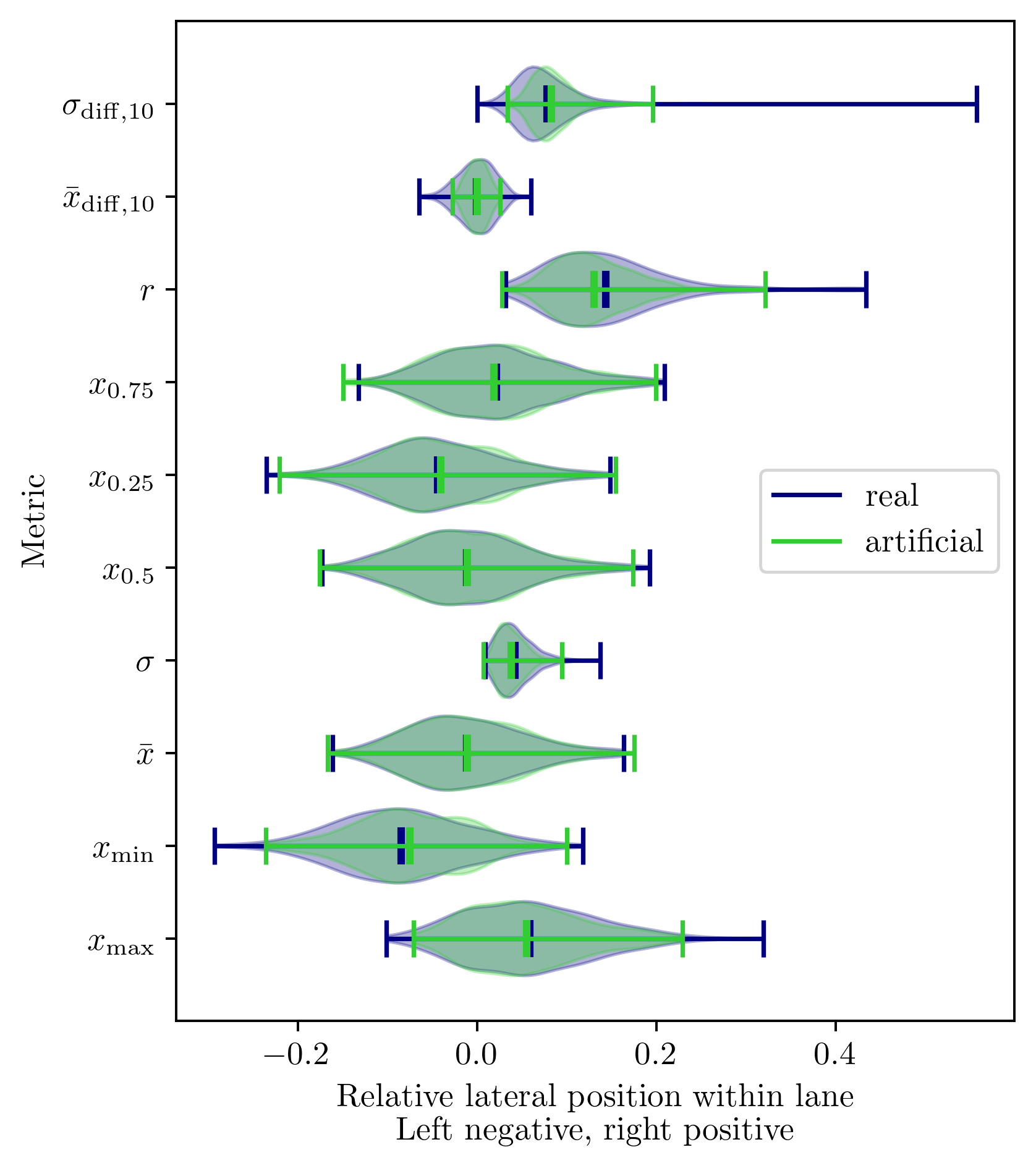}   
\caption{Comparison of metrics for real lateral offset profiles and lateral offset profiles generated by keeping the capped real fine movement and adding an artificial coarse movement created by the Markov model.}
\label{fig:coarse_art}
\end{figure}

In the next step, the quality of the Markov model is evaluated. The artificial lateral offset profiles are obtained by keeping the capped real, not shifted fine movement and replacing the real coarse lateral offset profile by an artificial one generated by the Markov model. The results are given in Fig.~\ref{fig:coarse_art}. Overall, the deviations from the real results for the metrics are only slightly greater than in Fig.~\ref{fig:shift} and there is still good agreement between both. 

\subsection{Evaluation of noise model}
\begin{figure}
\includegraphics[width=0.5\textwidth]{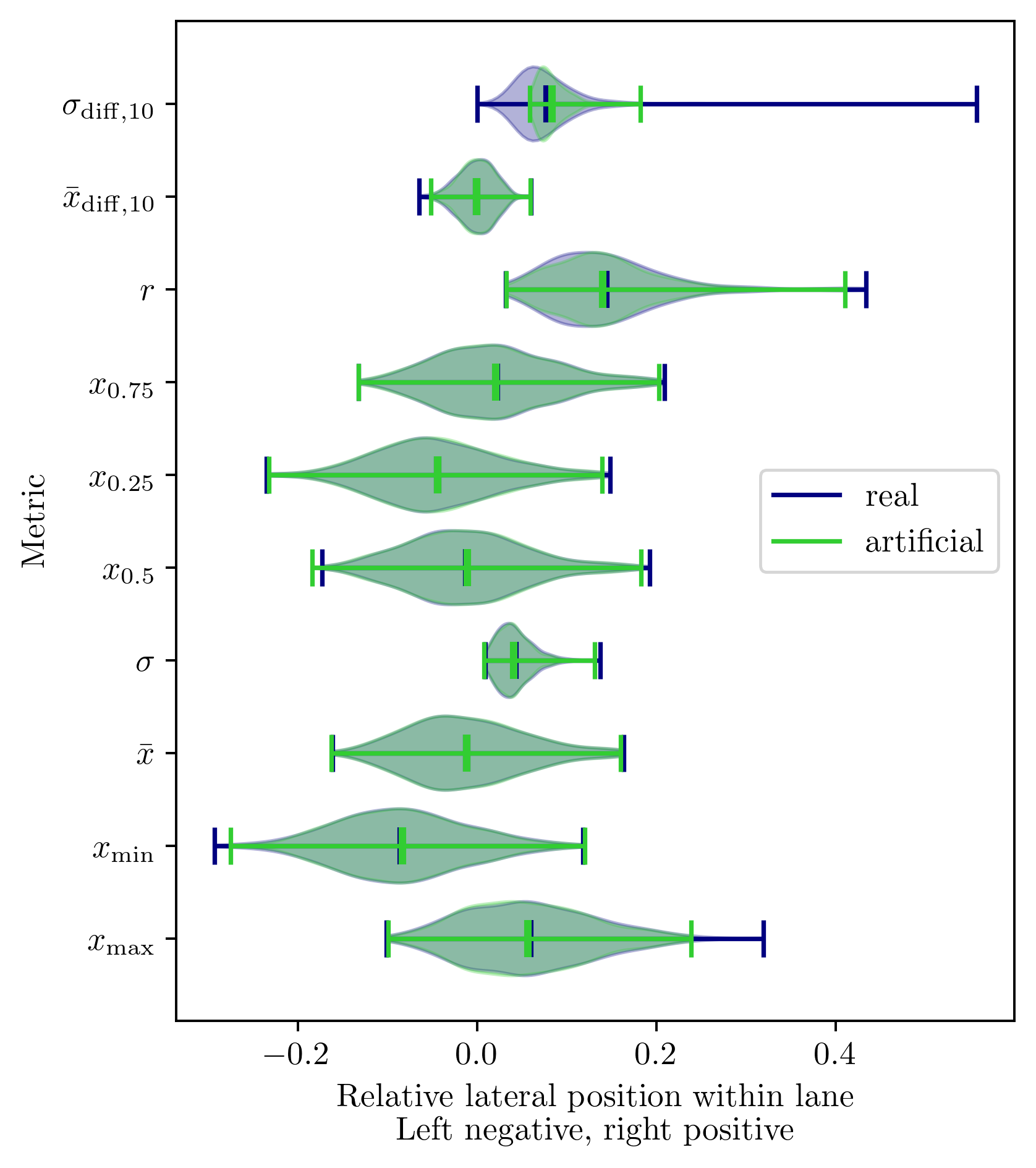} 
\caption{Comparison of metrics for real lateral offset profiles and lateral offset profiles generated by keeping the real coarse movement and adding an artificial fine movement created by the noise model.}
\label{fig:fine_art}
\end{figure}

Similarly to the previous step, this section evaluates the quality of the noise model only. The artificial lateral offset profiles consist of the real coarse lateral offset profile and added noise generated by the noise model. Fig.~\ref{fig:fine_art} illustrates the results. Besides the metric $\sigma_\mathrm{diff,10}$ the results are close to what has been achieved with the time-shifted capped real fine movement in Fig.~\ref{fig:shift}.

\subsection{Evaluation of full two-level stochastic model}
\begin{figure}
\includegraphics[width=0.5\textwidth]{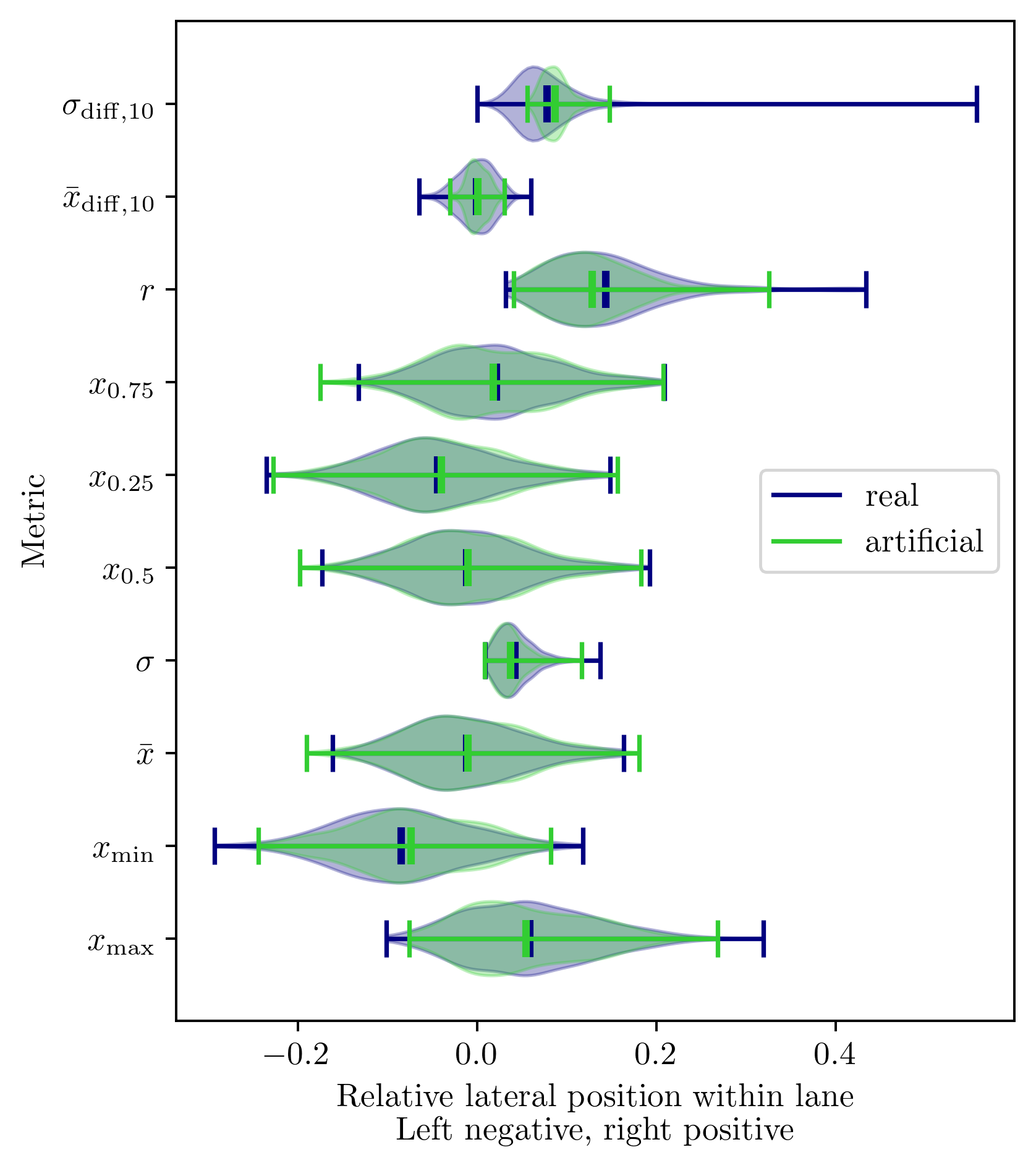}   
\caption{Comparison of metrics for real lateral offset profiles and lateral offset profiles generated by the combination of Markov and noise model.}
\label{fig:full_art}
\end{figure}

\begin{figure}
\begin{subfigure}{0.5\textwidth}
    \includegraphics[width=\textwidth]{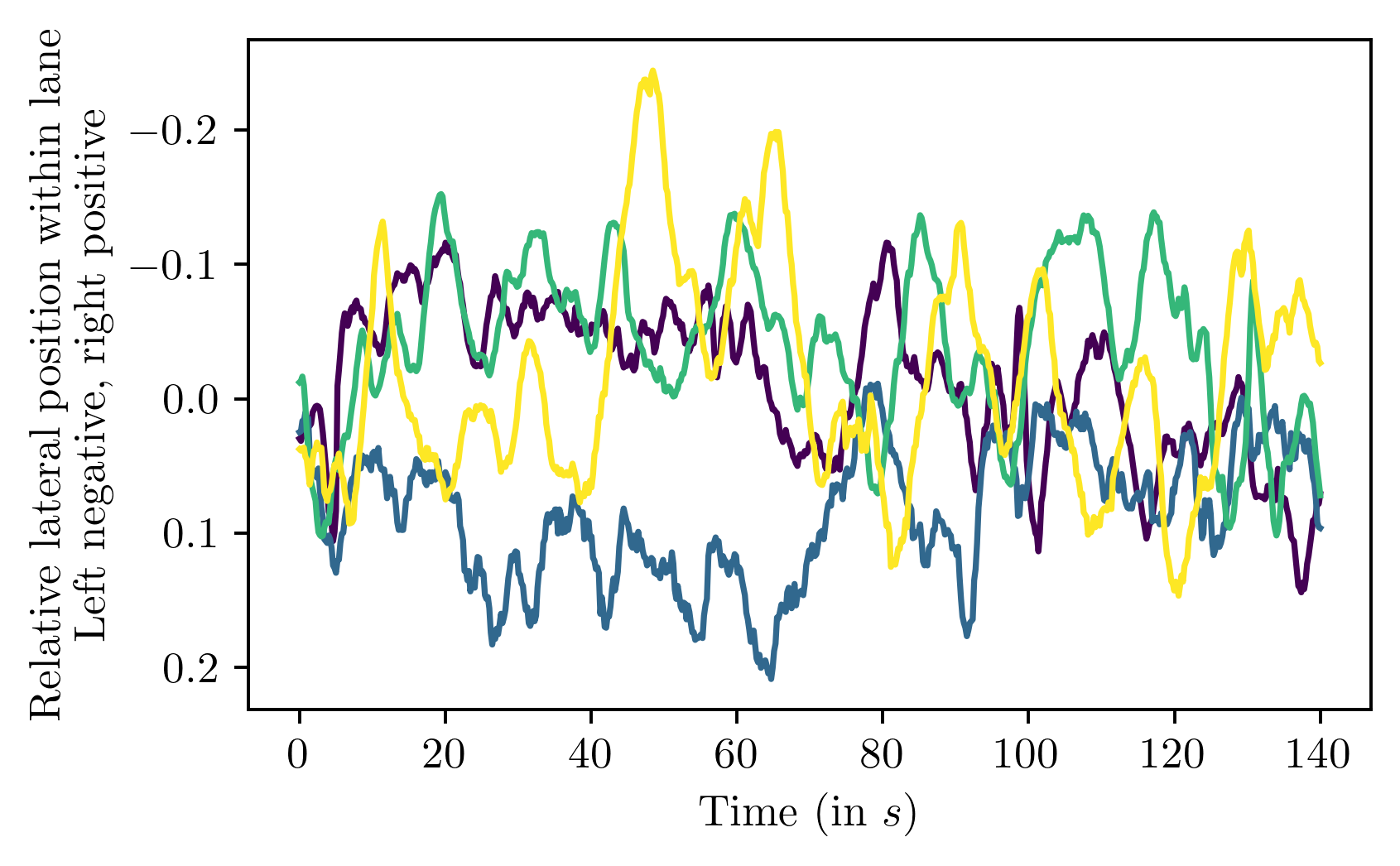}   
    \caption{Real lateral offset profiles}
    \label{fig:sample_orig_traj}
\end{subfigure}
\begin{subfigure}{0.5\textwidth}
    \includegraphics[width=\textwidth]{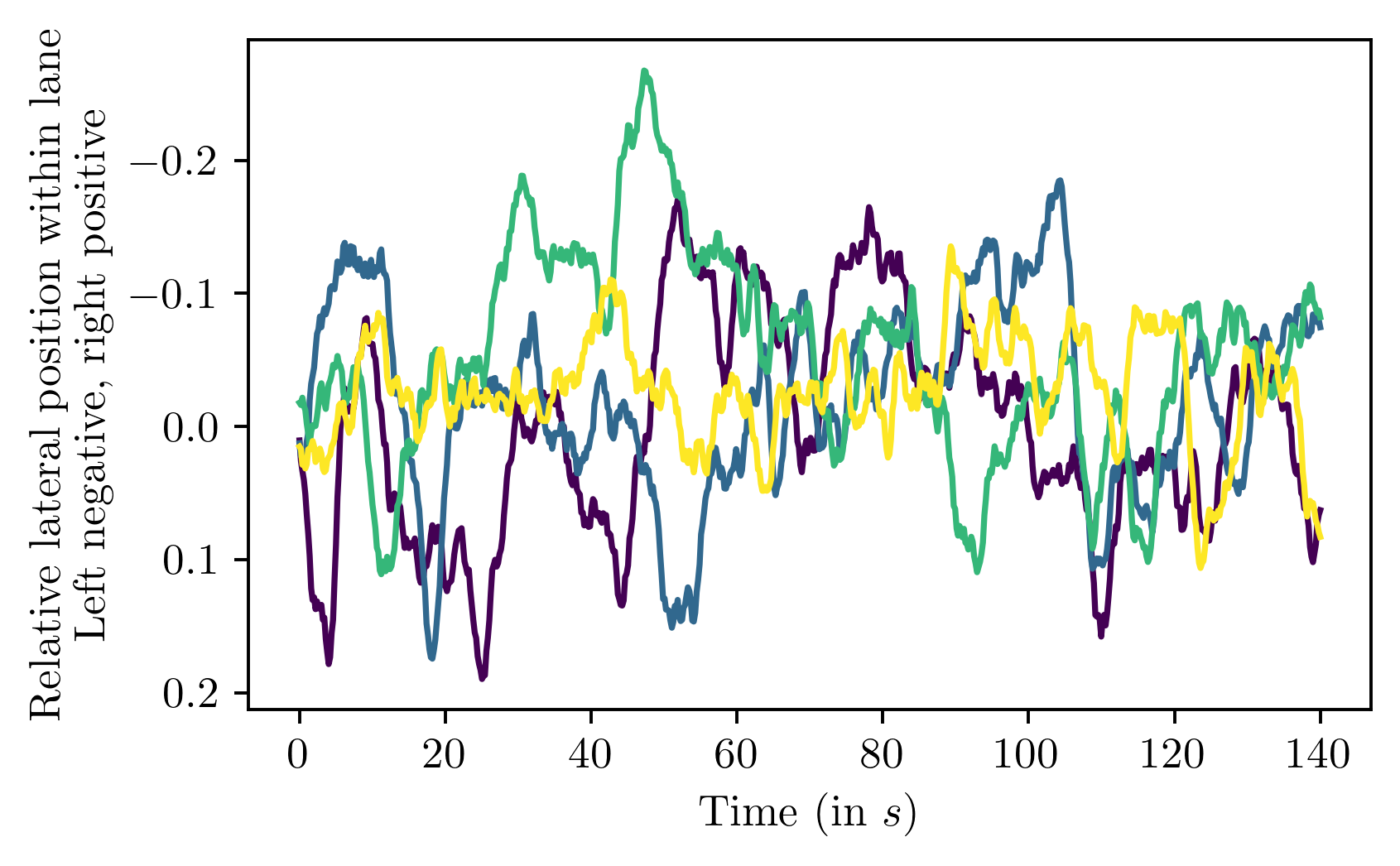} 
    \caption{Artificial lateral offset profiles}
    \label{fig:sample_art_traj}
\end{subfigure}
\caption{Illustration of real and artificial lateral offset profile generated by two-level stochastic model. Lateral offset profiles of the same color have the initial position in common.}
\label{fig:sample_traj}
\end{figure}

In the last step the performance of the full two-level stochastic model is assessed. For this, completely artificial lateral offset profiles are generated by combining the Markov model for the coarse movement and the noise model for the fine movement. The results obtained are visualized in Fig.~\ref{fig:full_art}. For $\bar{x}_\mathrm{diff,10}$ and especially $\sigma_\mathrm{diff,10}$ discrepancies between the real and the artificial lateral offset profiles can be noted. The plots for the remaining metrics show a good alignment.
Fig.~\ref{fig:sample_art_traj} illustrates sample artificial lateral offset profiles generated by starting from the initial lateral position of the real lateral offset profiles depicted in Fig.~\ref{fig:sample_orig_traj}. Even though the results for $\bar{x}_\mathrm{diff,10}$ and $\sigma_\mathrm{diff,10}$ indicate some weaknesses of the artificial lateral offset profiles, it is not possible to distinguish them from the real ones based on the plots. 

Considering the results of this and the previous sections it can be concluded that the Markov model is well suitable to model the coarse movement of a vehicle within its lane. Also the noise model shows good qualitative results and good agreement for eight out of the ten metrics. However, there is potential for improvement regarding the jumps of the lateral offset from one time step to the next.

\begin{figure}
\includegraphics[width=0.45\textwidth]{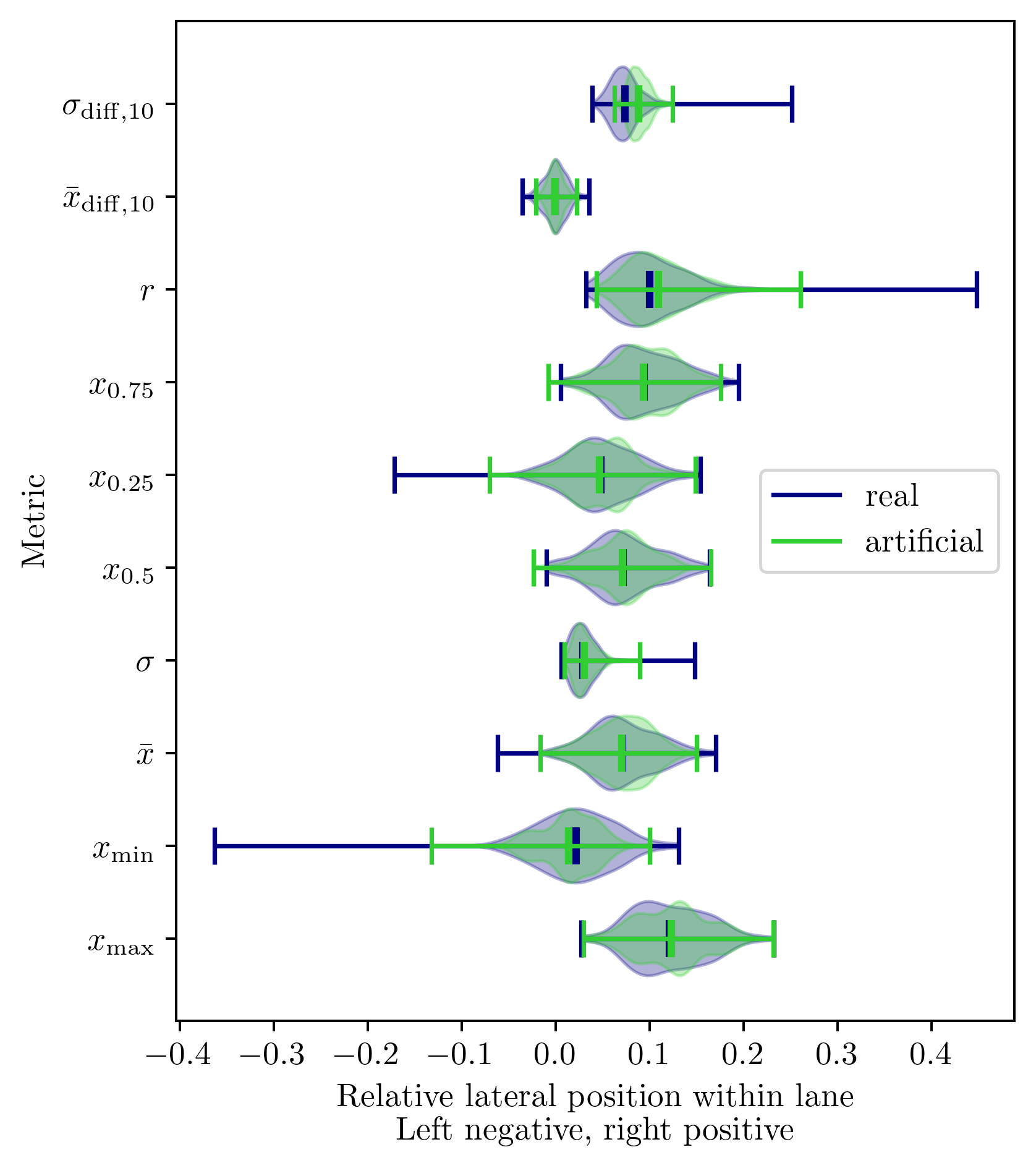}   
\caption{Comparison of metrics for real lateral offset profiles and lateral offset profiles generated by two-level stochastic model for a different tour and a different driver.}
\label{fig:diff_driver}
\end{figure}

To show how the model is able to capture the characteristics of different tours, it is applied to the second tour dataset. The results are illustrated in Fig.~\ref{fig:diff_driver}. It can be seen that the results obtained for the metrics using this drive differ from the ones seen before. With the given data, the differences cannot clearly be assigned to the driver as also other factors within the tour such as the traffic density might be responsible for those. However, it can be seen that the model is able to reflect such differences and thus it can be expected that it would also be able to capture driver-specific characteristics in particular. 

Another advantage of the introduced two-level stochastic model is its lightness: on a single core of a machine with an \mbox{Intel Core i7-10850H processor} running at \num{2.7}~\si{\giga\hertz}, \num{18000} evaluations of the model corresponding to the calculation of the lateral offset profile for a \num{1}~\si{\hour} drive took about \num{0.36}~\si{\second}. Thus, the model is about \num{10000} times faster than real-time. This makes it suitable for Hardware-in-the-loop and batch simulations. The fine movement generated by the noise model can even be calculated offline in advance, saving additional \num{5}~\% computation time. Note also that the model is computed per vehicle, i.e. in practical applications it will scale linearly w.r.t. the number of simulated vehicles.

\section{MODEL ENHANCEMENT}\label{sec:ModelEnhancement}
A limitation of the developed model is that it is only based on the overall lateral offset profile of a given tour. 
In order to create artificial lateral offset profiles that do not only represent the general characteristics but are realistic with respect to real-world behavior, the model has to be further enhanced. Existing literature and the analysis performed in Sec. \ref{sec:Velocity} reveal dependencies of the lateral movement on the static environment (e.g. the lane (left/center/right), the lane width \cite{wangImpactLaneWidth2014, dijksterhuisEffectsSteeringDemand2011}), the surrounding traffic participants (e.g. oncoming traffic \cite{dijksterhuisEffectsSteeringDemand2011} and the vehicle in front \cite{krajewskiHighDDatasetDrone2018}) and ego vehicle related aspects (e.g. longitudinal velocity as shown in Fig.~\ref{fig:velocity}). Including aspects such as the ego vehicle's longitudinal velocity into the model particularly implies the consideration of dependencies between the lateral and the longitudinal movement.
As the work performed within the scope of this paper indicates that the fine lateral movement is independent from the coarse lateral position of a vehicle, it can be assumed that the enumerated factors mainly affect the latter. Thus, in particular the Markov model needs to be enhanced. A possible approach is to extend it to a Hidden Markov model with hidden states representing influencing factors such as the lead vehicle. 

The Markov model used within this paper is a first order Markov model. The good performance of this models suggest the assumption that this is sufficient for the considered use case of modelling the general characteristics of a tour. However, when considering influences with a temporal extension such as the overtaking of a vehicle, more than one previous state might be relevant to determine the next one, requiring a higher level Markov model. 

As stated earlier, the noise model's performance regarding the metrics $\bar{x}_\mathrm{diff,10}$ and $\sigma_\mathrm{diff,10}$ is another candidate for advancement.

A further possible enhancement of the model is the consideration of driving physics. As long as the vehicle is operating far away from its physical boundaries, these can be handled by a downstream control model that physically implements the human behavior. However, when considering scenarios such as near misses in which a system comes close or reaches its physical boundaries, an interlocking of the behavioral and physical model might be necessary.  

\section{CONCLUSION}\label{sec:Conclusion}
Most simulations of homogeneous traffic conditions currently neglect or simplify the lateral movement of vehicles within their lane. However, as outlined, there is a clear necessity for such models. Therefore, this paper introduces a two-level stochastic model to describe the lateral movement of vehicles within their lane. It consists of a Markov model for a vehicle's coarse movement and a noise model for the independent fine movement. From the performed evaluation it can be seen that this split into two granularity layers can be rated permissible for the chosen model setup. While the results of the Markov model are in very good agreement with the real behavior, the noise model shows potential for improvements regarding two out of the ten defined metrics. The two-level stochastic model is able to capture characteristics of different tours. Thus, it can be expected that it is suitable to model individual drivers within a simulation. An additional advantage of the model is its extremely low computation time. 
Moreover, the model is flexible due to its formulation in terms of stochastic models with established generalizations, in particular the extension of Markov chains to Hidden Markov models. This supports the consideration of surrounding traffic interactions and ego vehicle related factors influencing the lateral offset profile in a next step. First potential candidate factors and their possible effects have already been identified.  

\bibliographystyle{ieeetr}
\bibliography{bibliography} 

\end{document}